\documentclass{article}

\usepackage{arxiv}

\usepackage[utf8]{inputenc} 
\usepackage[T1]{fontenc}    
\usepackage{hyperref}       
\usepackage{url}            
\usepackage{booktabs}       
\usepackage{amsfonts}       
\usepackage{nicefrac}       
\usepackage{microtype}      
\usepackage{lipsum}

\usepackage{hyperref}
\usepackage{csquotes}
\usepackage{graphicx}
\usepackage{amsmath}
\usepackage{bm}
\title{Performance Analysis of Semi-supervised Learning in the Small-data Regime using VAEs}

\author{
  Varun Mannam \thanks{Varun Mannam is with the Department of Electrical Engineering,
University of Notre Dame, Notre Dame, IN, 46556 USA e-mail:
vmannam@nd.edu} \\
  Department of Electrical Engineering\\
  University of Notre Dame\\
  Notre Dame, IN 46556 \\
  \texttt{vmannam@nd.edu} \\
   \And
 Arman Kazemi \\
  Department of Computer Science and Engineering\\
  University of Notre Dame\\
  Notre Dame, IN 46556 \\
   \texttt{akazemi@nd.edu}\\
}

\begin{document}
\maketitle

\begin{abstract}
Extracting large amounts of data from biological samples is not feasible due to radiation issues, and image processing in the small-data regime is one of the critical challenges when working with a limited amount of data. In this work, we applied an existing algorithm named Variational Auto Encoder (VAE) that pre-trains a latent space representation of the data that captures the features in a lower-dimension for the small-data regime input. The latent space representation will be fine-tuned, and its weights will be fixed. The latent space will be used as a segment of the neural network that can be used for classification. Here we will present the performance analysis of the VAE algorithm with various latent space sizes in the semi-supervised learning using the CIFAR-10 dataset.
\end{abstract}


\section{Introduction}

Artificial neural networks (ANNs), specifically Convolutional Neural Networks (CNNS), have become popular due to their success in image classification, feature extraction and object recognition and detection \cite{goodfellow2016deep} in the recent years. CNNs leverage the huge amount of labeled data available to train networks that outperform humans in image recognition tasks. However, in the small-data regime, the accuracy of trained networks using a limited number of labeled samples is low \cite{chawla2005learning}. This is a typical case when working with biological samples where exposure to radiation (in order to capture an image) is detrimental to the well-being of the sample. More images can be derived from the initial data by some augmentation methods, but it is unhelpful due to lack of labeled images. \newline

To address this problem, there exists a framework called \enquote{Auto Encoder} (AE \cite{wikiautoencode}) that uses all the input data, labeled and unlabeled, to train a low-dimensional embedding. AE is a neural network that takes unlabeled images as the input and regards the input itself as the label. As illustrated in Figure \ref{fig:1}, AE is comprised of two parts: the encoder and the decoder. The encoder part tries to embed the features in a latent space that can extract the features of the original image and the decoder tries to restore the image to the original image. The process called \enquote{pre-training} trains the weights for both encoder and decoder parts. Once trained, the encoder part of the AE will be a representation of all the labeled and unlabeled data. This increases the amount of usable information from all of the images. \newline

Traditional semi-supervised models consists of pretraining using Restricted Boltzmann machine (RBM)\cite{Hinton} or Gaussian-Restricted Boltzmann machine (G-RBM)\cite{Hinton}. RBM is an energy based model that is represented using an undirected graph containing a layer of observable variables and a single layer of latent variables (similar to hidden units in a multi-layer perceptron) \cite{goodfellow2016deep,Murphy}. This energy based model was first introduced in 1980's \cite{Smolensky} and have been implemented using diverse datasets including image \cite{Hinton} and medical data \cite{Nguyen}.  Hinton and Salakhutdinov \cite{Salakhutdinov} showed that RBMs can be stacked and trained in a greedy manner. This deep learning model, Deep belief networks(DBN), that utilizes RBM as the learning model has been implemented on various unsupervised and supervised learning problems. Later, Bengio et. al. \cite{Bengio} showed that the pre-trained undirected graphical model in semi-supervised setting performs well with deep architectures. However, the challenge working with RBM is that it is constructed using sigmoid functions as the activation function between the input and the hidden layer. As we are aware that the major drawback of using sigmoid activation function is the vanishing gradient problem. Hence, in this work, we pre-train the model using \enquote{Variational Auto Encoder} (VAE \cite{wikiautoencode}).

After pre-training the encoder is able to observe similar images to the training images and extract the valuable features from it in a lower dimension than the initial image. Now it is possible to couple the encoder with a small neural network and train that network for classification tasks. The present work is similar to reinforcement learning where the model is trained with one dataset and uses the feature extraction part of that model to train another model for a different dataset. It is important to mention that the encoder weights are fixed and can't be changed. It is only the small neural network that will be trained. The input to this network is the small set of labeled data. This is called \enquote{fine-tuning}.

In this project we  will implement VAE  that tries to capture not only the compressed representation of the images, but also the parameters of a probability distribution representing the data. We will examine the effect of different size of the latent spaces and how it affects the accuracy of the model. Later, we will analyze the performance of the semi-supervised model with the optimum latent space.

\section{Background}
In this section we enumerate the basic details about AE and VAE.  
\subsection{Auto Encoder}
An autoencoder is a type of ANN used to learn efficient data encoding in an unsupervised manner. The aim of an autoencoder is to learn a representation of a set of data, typically for dimensionality reduction, by training the network to ignore signal noise. Along with the reduction side, a reconstructing side is learned, where the autoencoder tries to generate from the reduced encoding a representation as close as possible to its original input. 
An autoencoder always consists of two parts, the encoder and the decoder, which can be defined as transitions $\phi$ and $\psi$ such that \cite{wikiautoencode}:
\begin{equation}
\phi:X \rightarrow F 
\end{equation}
\begin{equation}
\psi:F \rightarrow Y \newline
\end{equation}
\begin{equation}
\phi, \psi = argmin_{\phi, \psi} ||X - (\psi  o  \phi) (X)^2||
\end{equation}
where the given input is $X$ and the predicted output is $Y$. If the feature space $F$ has lower dimensionality than the input space $X$, then the feature vector $\phi (x)$ can be regarded as a compressed representation of the input $X$.

\begin{figure}[!h]
    \centering
    \includegraphics[width=0.8\textwidth]{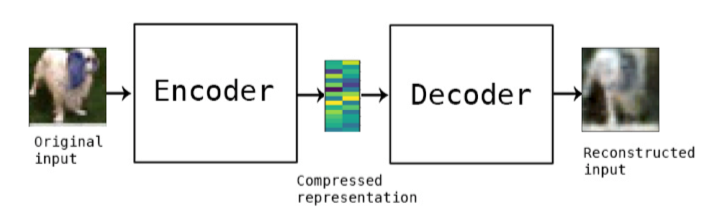}
    \caption{Autoencoder block diagram} 
    \label{fig:1}
\end{figure} 

\subsection{VAE}
To extend the idea used in the AE, there is a variation called VAE which uses the ``KL-divergence'' between the predicted probability distribution function and actual posterior distribution in the latent space whereas traditional AEs try to find the accurate mapping functions in the encoder (between input and latent space) as well as in the decoder (between the latent space and output). Using VAE, we generate a large dataset by adding the noise in the latent space which is similar to input data augmentation (adding noise to images to increase the number of examples in the input dataset). 

VAE uses a variational approach for latent representation learning, which results in an additional loss component. It assumes that the data is generated by a directed graphical model $p(\textbf{X}|\textbf{Z})$ and that the encoder is learning an approximation $q_{\phi}(\textbf{Z}|\textbf{X})$ of the posterior distribution $p_{\theta}(\textbf{X}|\textbf{Z})$ where $\phi$ and $\theta$ denote the parameters of the encoder (recognition model) and decoder (generative model) respectively. We can write the conditional or posterior distribution 
\begin{equation}
    p(\bm{z}|\bm{x}) = \frac{p(\bm{z},\bm{x})}{p(\bm{x})} = \frac{p(\bm{x}|\bm{z}) p(\bm{z})}{p(\bm{x})}
\end{equation}
The  denominator of above equation is the marginal distribution of the observations and is calculated by marginalizing out the latent variables from the joint distribution, i.e.
\begin{equation}
    p(x) = \int_{z} p(z,x)dz
\end{equation}
In  many  cases  of  interest  this  integral  is  not  available  in  closed  form  or  is  intractable (requires exponential time to compute). Hence, we consider variational approximation as follows: consider a tractable distribution q(z).  The goal is  to  find  the  best  approximation,  e.g.,  the  one  that  satisfies  the  following  optimization
problem:
\begin{equation}
    Minimize: D_{KL}[q_{\phi}(\bm{z}|\bm{x})|| p_{\theta}(\bm{z}|\bm{x})]
\end{equation}
Therefore, the objective of the variational autoencoder in this case has the following form: 
\begin{equation}  \label{loss}
\mathcal{L}(\phi, \theta, \mathbf{x})=D_{\mathrm{KL}}\left(q_{\phi}(\mathbf{z} | \mathbf{x}) \| p_{\theta}(\mathbf{z})\right)-\mathbb{E}_{q_{\phi}(\mathbf{z} | \mathbf{x})}\left(\log p_{\theta}(\mathbf{x} | \mathbf{z})\right) 
\end{equation} where $D_{KL}$ stands for the KL-divergence. 
In the VAE, the principle is to minimize the loss between input and the restored image along with the loss generated by the latent space to represent the features in the input images. 

\section{Methodology}

\begin{figure}
    \centering
    \includegraphics[width=0.5\textwidth]{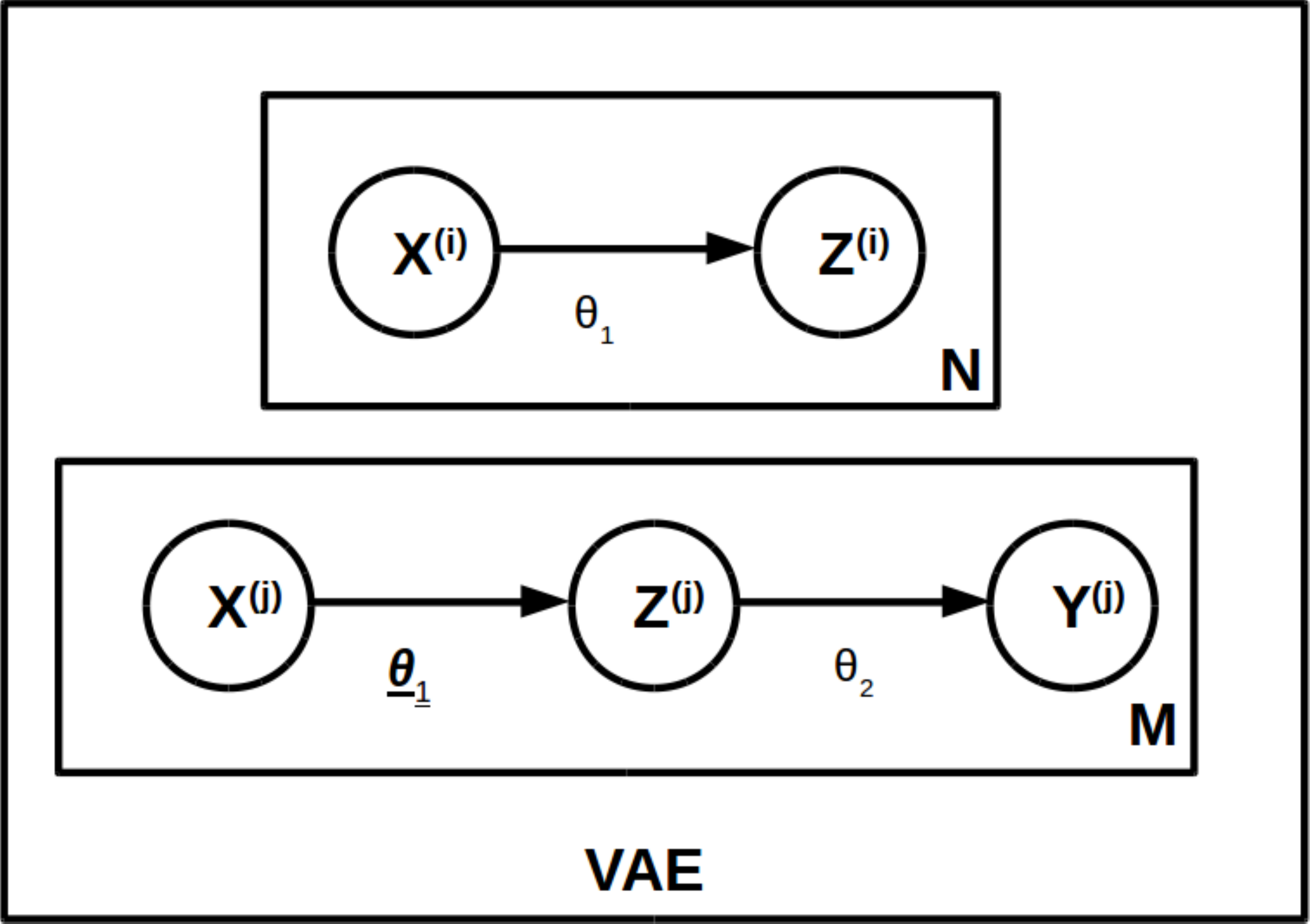}
    \caption{Network architecture: pre-training with VAE(first row) In the second row, parameter $\theta$ is underlined and bold to indicate that these parameters are freezed when the fine-tuning the network}
    \label{fig:my_label}
\end{figure}

Consider a surrogate model $y = f(x,\theta)$ which is trained using limited simulation data $\mathcal{D} = \{x^i,y^i\}_{i=1}^{N},\{x^j\}_{j=N+1}^{D}$. Where the input data, $x^i\in \mathbb{R}^{d_x\times H \times W}$ is the input from CIFAR-10 dataset. Here H and W are the height and width respectively and $d_x$ is the number of dimensions for the input x at one location.  $x^j$ is the additional data utilized for pretraining the model. $y^i \in \mathbb{R}^{1}$  is the classified result. $\theta$ is the model parameter and N is the total number of training data utilized during fine-tuning and D is the total number of data utilized for pre-training. In semi-supervised model, we pre-train the model with the  input data $\mathbb{R}^{d_x\times H \times W}$ and then perform image classification problem $\mathbb{R}^{d_x\times H \times W} \rightarrow \mathbb{R}^{ 1}$.
For both the pre-training and fine tuning, we used stochastic gradient descent with Adam optimizer to update the network weights and biases. The simulation was performed using Pytorch machine learning package in Python. 
\subsection{VAE pre-training}
We implemented dense-net \cite{huang2017densely},\cite{ zhang2018poisson} version of VAE for the pre-training part. Dense-net contains the encoder and decoder blocks along with the dense-layer that has the simple and complex features.

\subsection{VAE fine-tuning}
We implemented a simple fully connected layer to classify the input images on the CIFAR-10 \cite{krizhevsky2009learning} data. This is due to the expectation that the latent space is smaller (either 4 $\times$ 4 or 8 $\times$ 8) (image size is 32 $\times$ 32). If the number of channels are large at the latent space, we will add more fully connected layers for classification of images.

\section{Data}
We perform the simulations on the CIFAR-10 dataset with ten image classes with three input channels $(C=3)$ of size 32 $\times$ 32 (W $\times$ H). CIFAR-10 dataset has 50000 training images and 10000 test images. 
\section{Results}
In this section we enumerate the results obtained using CIFAR-10 data. We consider the following latent dimensions: 6400, 10,000 and 14,400. In order to evaluate the performance of the model for above three latent space, we consider the distribution estimated for the values at various pixel location. 
\subsection{Pre-training}
For the results presented in this section, we have a dataset with 50,000 $\{x^k\}_{k=1}^{50000}$  examples for pre-training and the test set consist of 10000, $\{x^k\}_{k=1}^{10000}$ examples. Adam optimizer was used for training 100 epochs, with learning rate of 1e-4 and a plateau scheduler on the test RMSE. Batch size is always smaller than the number of training data. In this work, a batch size of 16 for pre-training was used. Weight decay was set to 1e-3 for pre-training.
\\

We consider Equation: \ref{loss} loss function to evaluate the trained model on test data and also to monitor the convergence.  
\begin{figure}[]
    \centering
    \includegraphics[width=0.5\textwidth]{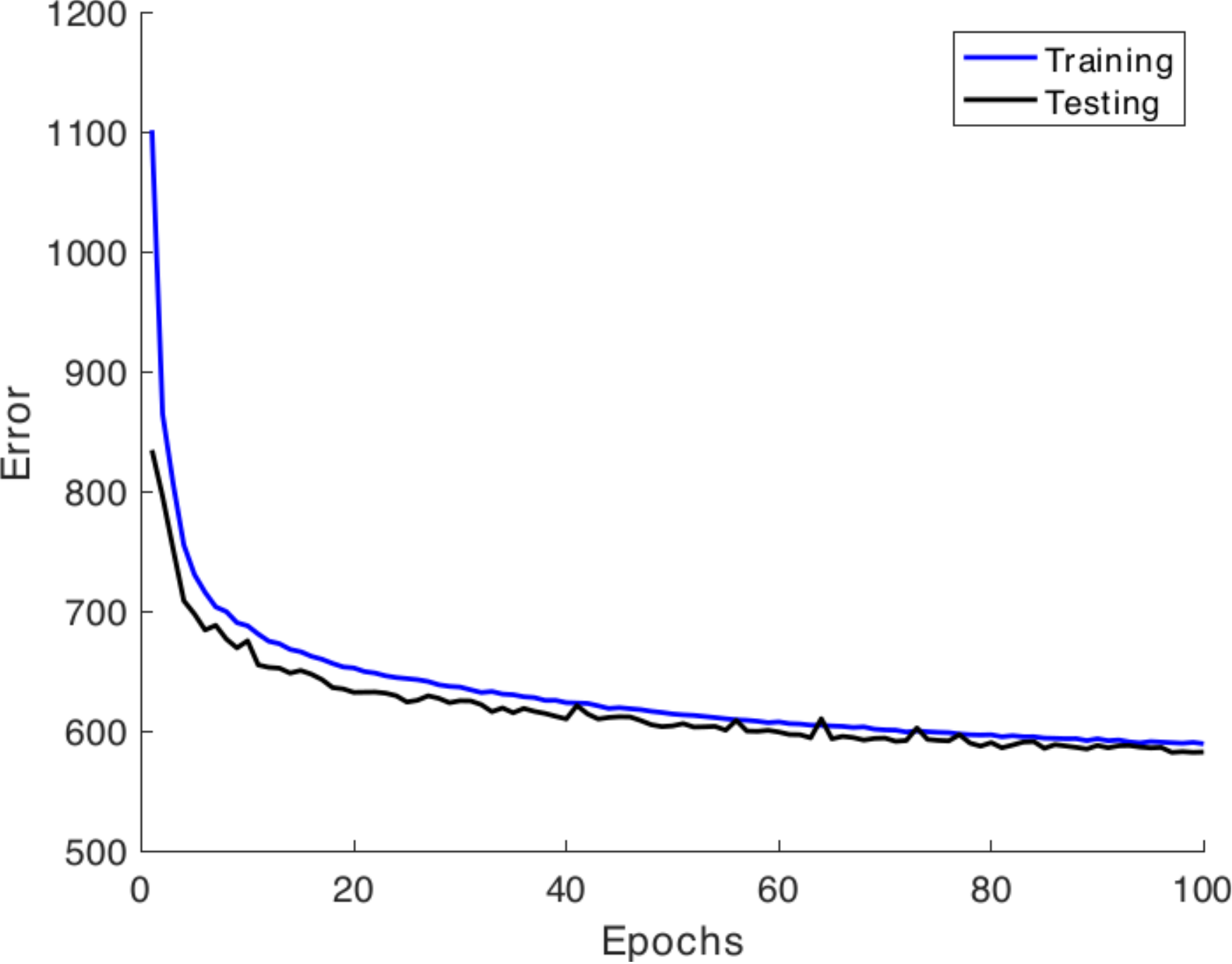}
    \includegraphics[width=0.5\textwidth]{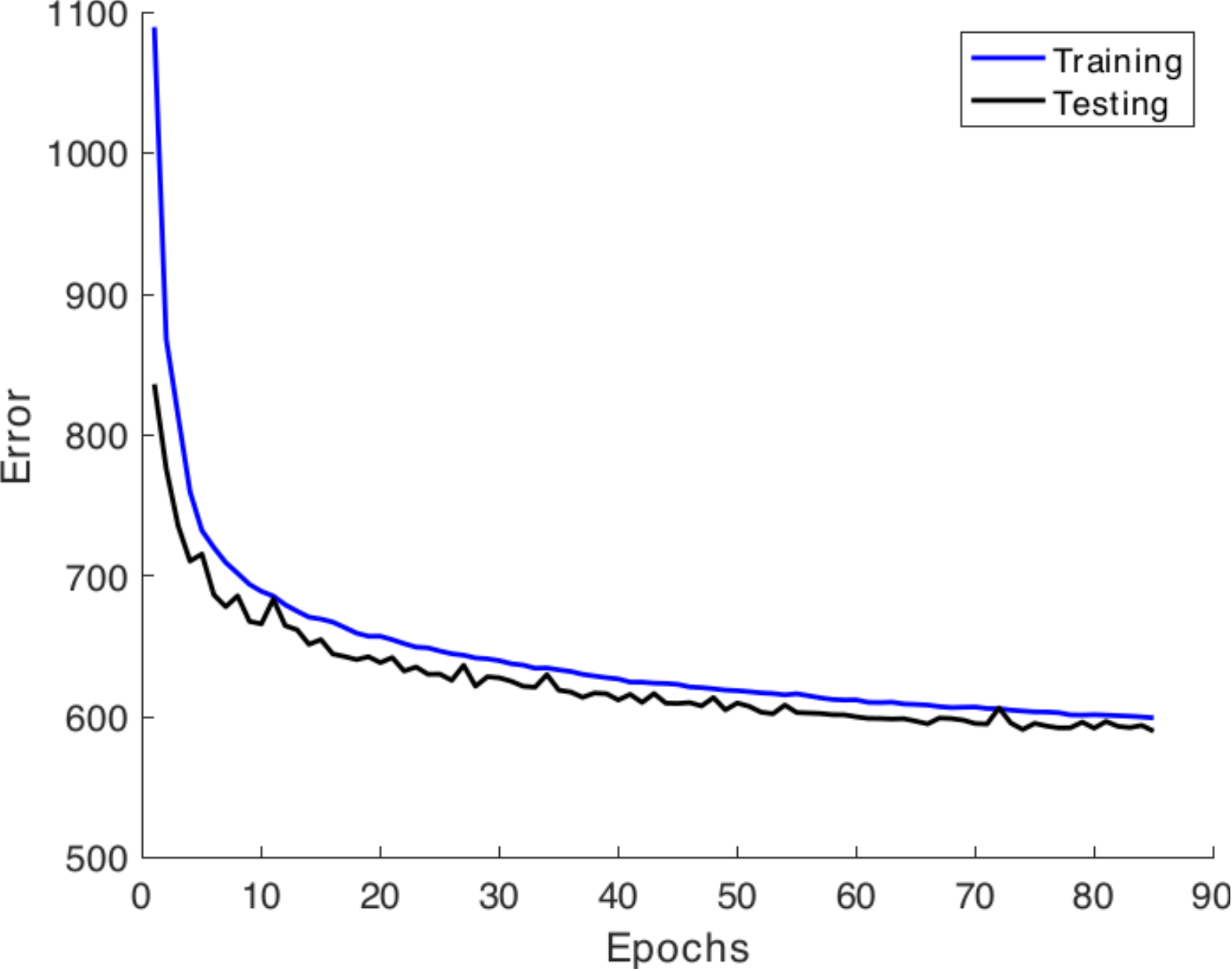}
    \includegraphics[width=0.5\textwidth]{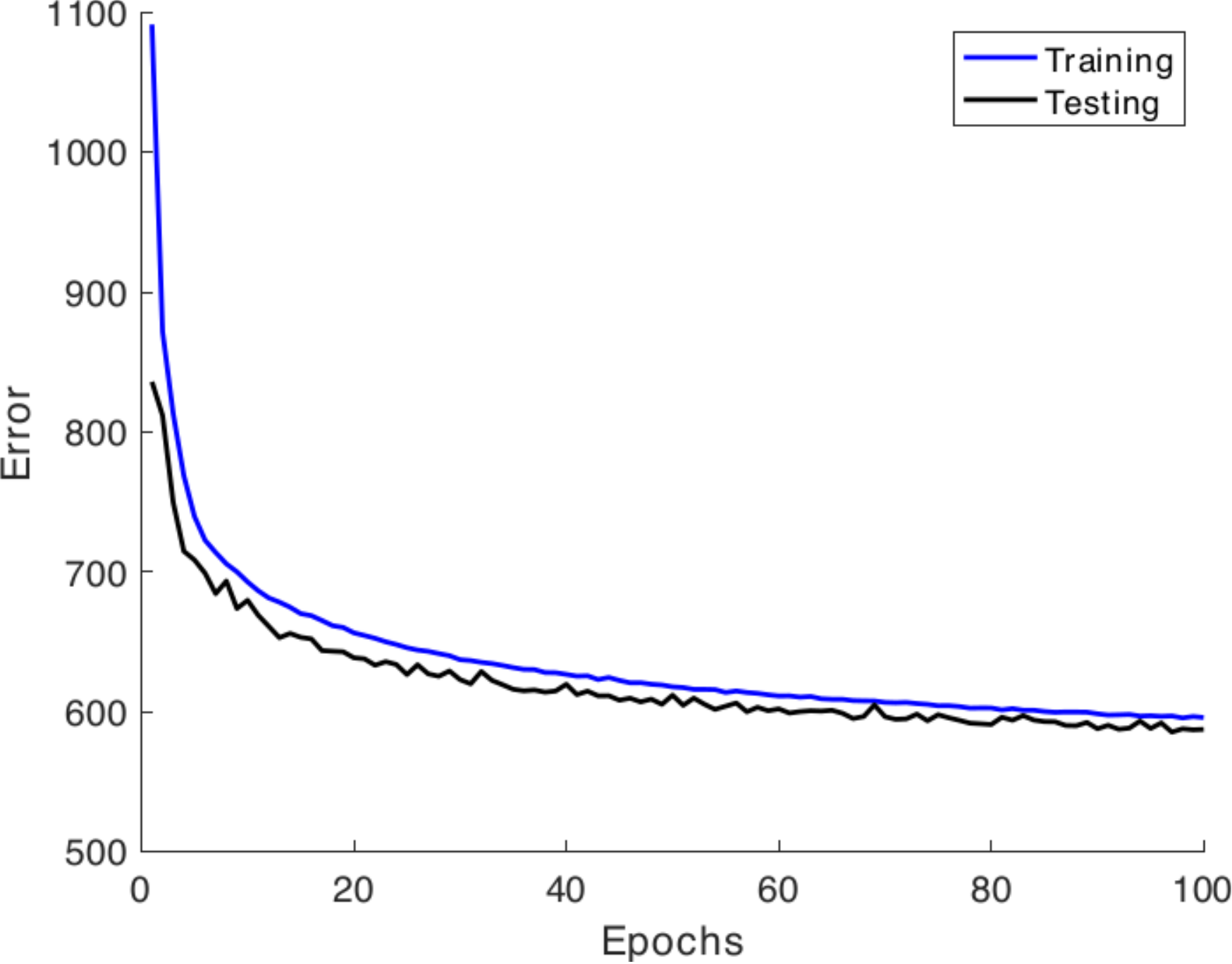}
    \caption{Error v/s epoch for 6400 latent space (top),  Error v/s epoch for 10000 latent space (middle) and Error v/s epoch for 14400 latent space (bottom)} \label{fig:3}
\end{figure} 

From figure \ref{fig:1}, we observe that the solution is converged after 50 epochs and most importantly the loss for the three latent spaces is similar.

\begin{figure}[]
    \centering
    \includegraphics[width=0.45\textwidth]{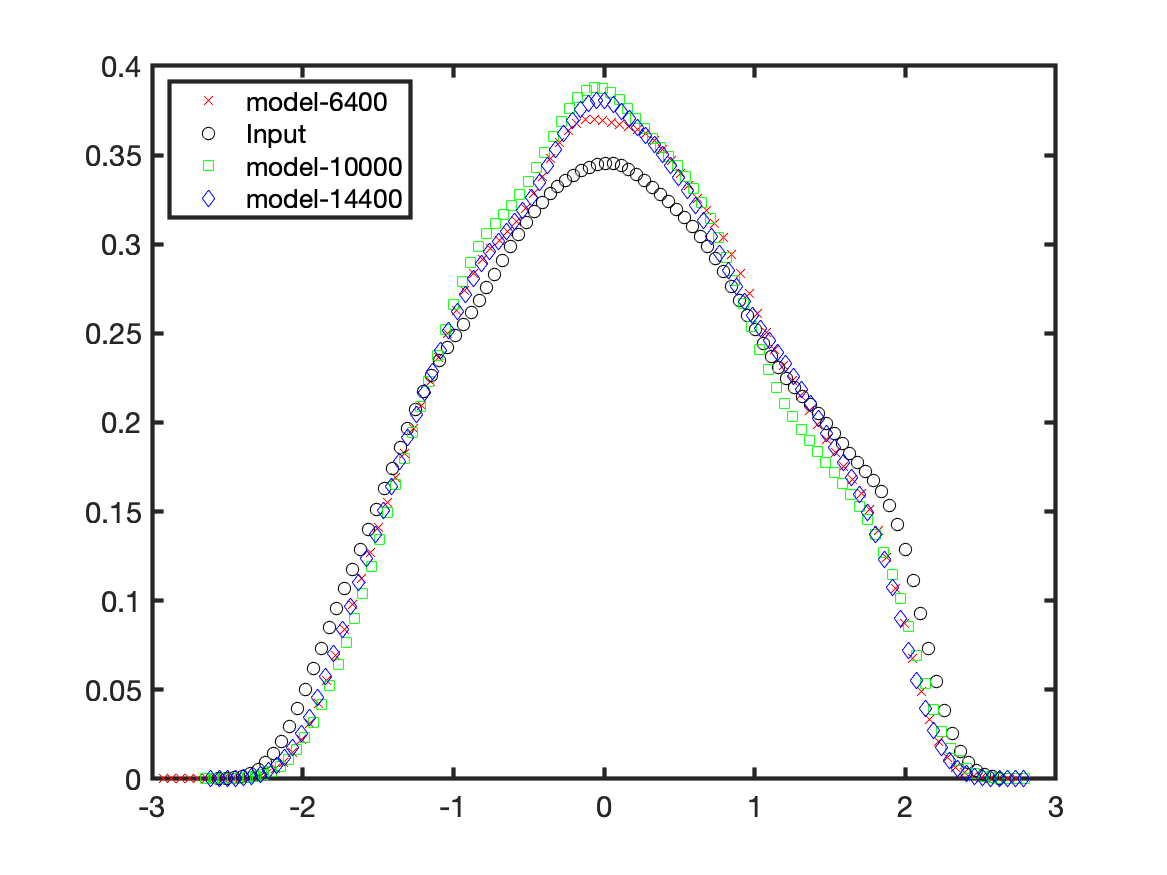}
    \includegraphics[width=0.45\textwidth]{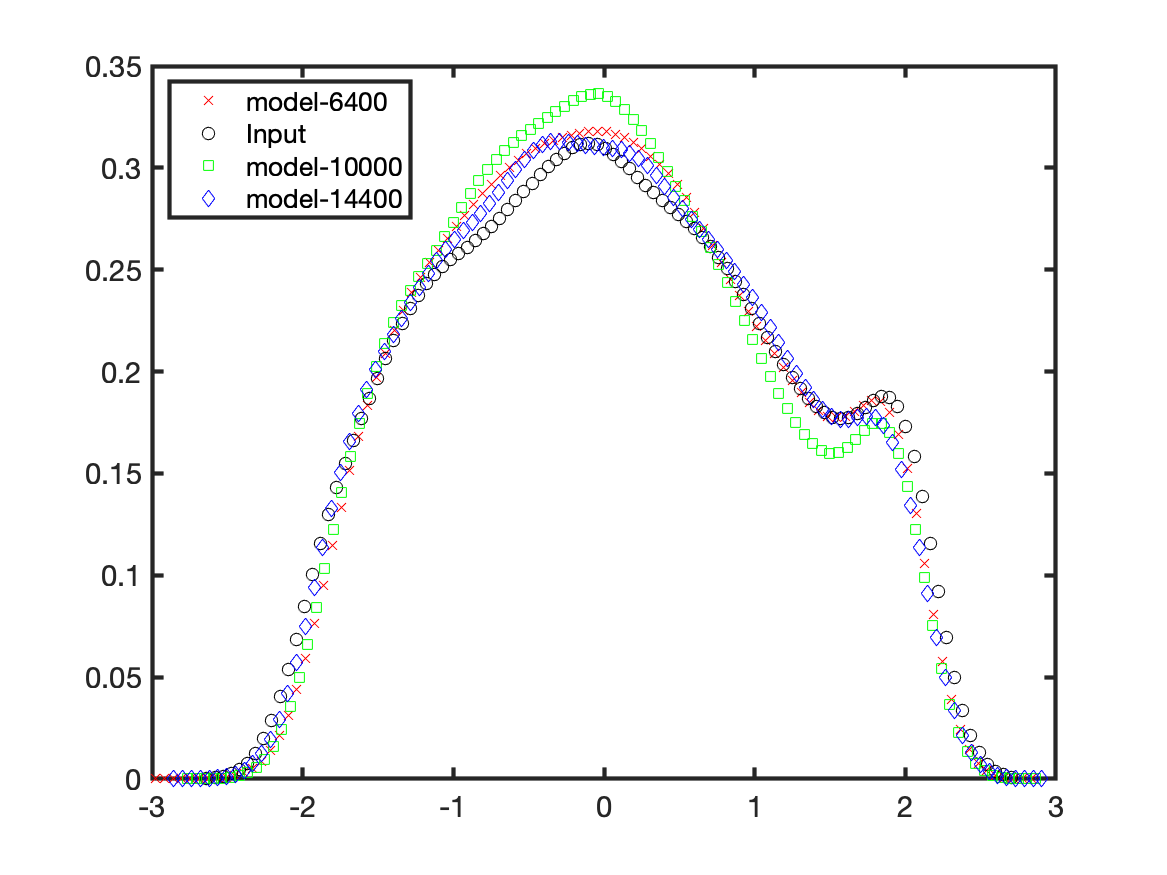}
    \includegraphics[width=0.45\textwidth]{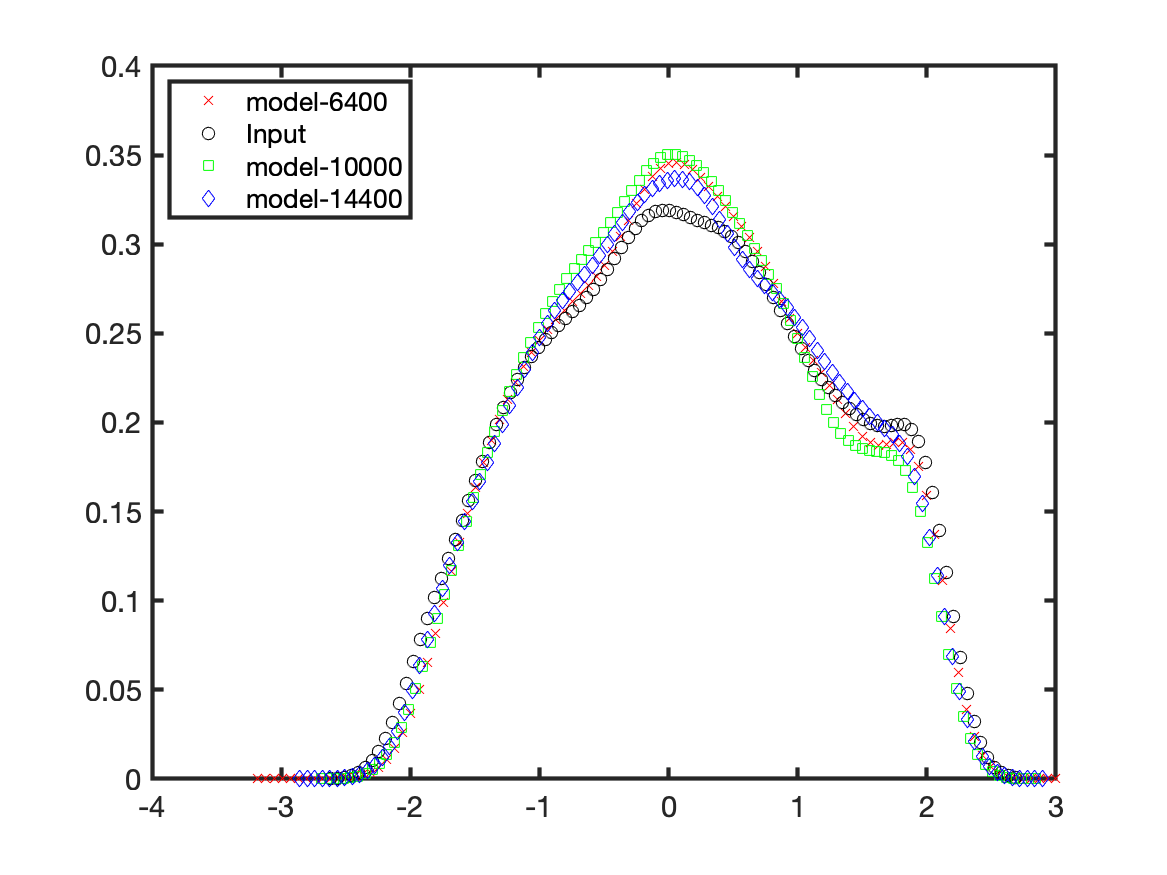}
    \includegraphics[width=0.45\textwidth]{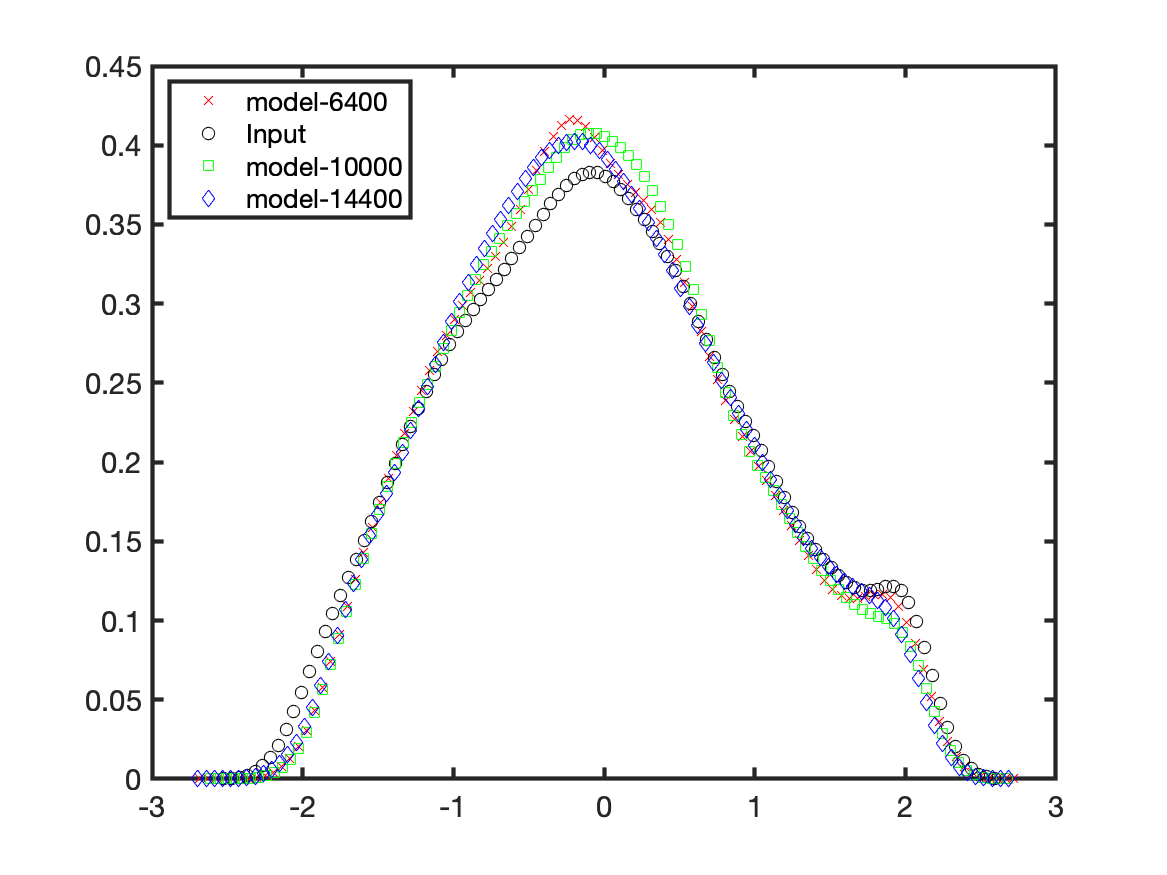}
    \caption{Distribution estimate for the values at various location of the square domain for 6400, 10000 and 14400 latent space} 
    \label{fig:2}
\end{figure}

From figure \ref{fig:2}, we observe that even when the latent size is small (Batch $\times$ 100 (channels) $\times$ 8 $\times$ 8) and (Batch $\times$ 100 (channels) $\times$ 10 $\times$ 10) the reconstructed density estimate is close to actual input data. The PDF with the latent size 10000 is closer to the actual input and also 6400 \& 14400 latent space. Since, all the latent spaces yield the similar outputs,we fine-tune and compare the classification accuracy in the next section.
\subsection{Fine-tuning}
In this section we freeze the parameters (weights and bias) used in the pre-training stage and fine tune the parameters (weights and bias) in the classification network. For this problem, we consider small data from the given CIFAR-10 dataset and use fully connected layers to perform classification. Cross entropy loss function is commonly used for all classification problems, we implemented cross entropy to measures the performance of a classification mode. 
\begin{figure}
    \centering
    \includegraphics[width=0.75\textwidth]{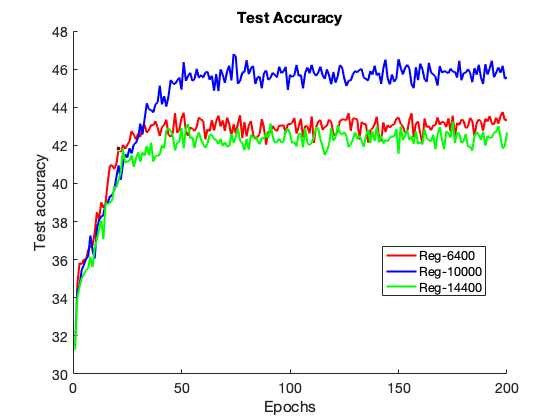}
    \caption{Fine tuning results with three different latent space models}
    \label{fig:reg}
\end{figure}
From figure \ref{fig:reg}, we observe the latent space 10000 yields better accuracy than other two latent spaces (6400 and 14400). The smaller the latent space the lower the test accuracy, this is due to insufficient features to classify the data. Also, for the large latent space, the test accuracy is low, this is due to model complexity \cite{goodfellow2016deep}.



\section{Conclusions and Future work}
The present document outlines the development of surrogate model for semi-supervised problem. In this work, we have implemented VAE as a pre-training model and a feed forward deep learning model for the classification. The results obtained for differently sized latent spaces are presented. It was observed that there is a slight improvement in the  test accuracy when the latent space is 10000 in comparison with latent space of 6400 and 14400.

For future work, a Bayesian approach can be explored. Due to a limited amount of data, it is necessary to model appropriate surrogate, since it is important to quantify the epistemic uncertainty induced by limited data \cite{per}, \cite{Yinhao} and hence, a Bayesian probabilistic approach is a natural way of addressing this challenge.


\bibliographystyle{unsrt}  
\bibliography{vae_arxiv}  

\end{document}